\documentclass[10pt,twocolumn,letterpaper]{article}

\usepackage{cvpr}      

\usepackage{multirow}
\usepackage[pagebackref,breaklinks,colorlinks]{hyperref}

\def\paperID{11989} 
\def\confName{CVPR}
\def\confYear{2025}

\title{Fine Tuning without Catastrophic Forgetting via Selective Low Rank Adaptation}

\author{
    Reza Akbarian Bafghi$^{1,2}$ \quad
    Carden Bagwell$^{2}$ \quad
    Avinash Ravichandran$^{2}$ \\
    Ashish Shrivastava$^{2}$ \quad
    Maziar Raissi$^{3}$ \\[0.5em]
    {\tt\small reza.akbarianbafghi@colorado.edu} \\
    {\tt\small \{carden.bagwell, avinash.ravichandran, ashish.shrivastava\}@getcruise.com} \\
    {\tt\small maziar.raissi@ucr.edu} \\[1em]
    $^{1}$University of Colorado, Boulder \quad
    $^{2}$Cruise, LLC \quad
    $^{3}$University of California, Riverside
}

\begin{document}
\maketitle
\begin{abstract} Adapting deep learning models to new domains often requires computationally intensive retraining and risks catastrophic forgetting. While fine-tuning enables domain-specific adaptation, it can reduce robustness to distribution shifts, impacting out-of-distribution (OOD) performance. Pre-trained zero-shot models like CLIP offer strong generalization but may suffer degraded robustness after fine-tuning. Building on Task Adaptive Parameter Sharing (TAPS), we propose a simple yet effective extension as a parameter-efficient fine-tuning (PEFT) method, using an indicator function to selectively activate Low-Rank Adaptation (LoRA) blocks. Our approach minimizes knowledge loss, retains its generalization strengths under domain shifts, and significantly reduces computational costs compared to traditional fine-tuning. We demonstrate that effective fine-tuning can be achieved with as few as 5\% of active blocks, substantially improving efficiency. Evaluations on pre-trained models such as CLIP and DINO-ViT demonstrate our method's broad applicability and effectiveness in maintaining performance and knowledge retention.
\end{abstract}

\section{Introduction}
Deep learning models, traditionally trained on extensive datasets, are renowned for their ability to generalize across diverse classes and applications~\cite{Dosovitskiy2020AnII,He2015DeepRL,Tu2024ACL}. However, adapting these models to emerging domains—such as recognizing novel objects~\cite{Zheng2023PreventingZT} or specialized environments like autonomous driving—requires significant retraining and risks erasing prior knowledge, a phenomenon known as catastrophic forgetting~\cite{vandeVen2024ContinualLA}. Although fine-tuning offers a feasible approach for domain-specific adaptation, it often compromises model robustness against distribution shifts~\cite{Wortsman2021RobustFO,Kumar2022FineTuningCD}, reducing reliability in unfamiliar scenarios. This limitation has highlighted the importance of pre-trained, zero-shot models, such as CLIP~\cite{Radford2021LearningTV}, which excel in out-of-distribution (OOD) generalization and are capable of extending to new tasks with minimal additional training. However, fine-tuning for task-specific gains frequently leads to a degradation in both robustness and zero-shot capabilities~\cite{Han2024AnchorbasedRF}, underscoring the ongoing need for techniques that balance domain-specific accuracy with consistent generalization.

\begin{figure}
     \centering
     \includegraphics[width=\linewidth]{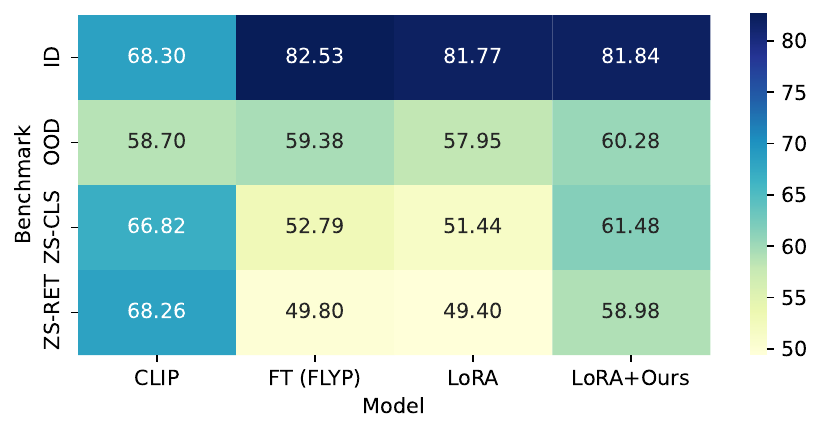}
    \caption{Comparison of pretrained CLIP, fully fine-tuned CLIP (FLYP), and LoRA fine-tuning with and without our method (rank 128) on ImageNet-1K (ID). Fine-tuning with FLYP and LoRA results in an average drop of 16.24\% and 17.12\% in zero-shot task performance (ZS-RET and ZS-CLS), respectively. In contrast, our method activates only 6.25\% of the LoRA blocks, reducing catastrophic forgetting by an average of 7.31\%, enhancing out-of-distribution (OOD) robustness, and improving inference efficiency. See Section~\ref{sec:vision-language}.}
    
     \label{fig:intro}
\end{figure}

Parameter-efficient fine-tuning (PEFT) has emerged as a promising strategy for balancing computational efficiency with knowledge retention. Initially developed to address the high computational demands of large language models (LLMs)~\cite{Hu2021LoRALA,Liu2024DoRAWL}, PEFT methods have shown potential for adapting vision and vision-language models~\cite{Zanella2024LowRankFA}. By updating only a subset of parameters, PEFT reduces the need for full model retraining, making domain adaptation feasible without prohibitive computational costs~\cite{Hu2021LoRALA}. Recent studies also indicate that PEFT can more effectively mitigate catastrophic forgetting than traditional fine-tuning~\cite{Biderman2024LoRALL,AkbarianBafghi2024ParameterEF}.

On the other hand, existing solutions, including Task Adaptive Parameter Sharing (TAPS)~\cite{Wallingford2022TaskAP} introduced an effective solution for tuning a pre-trained model for multiple tasks, using task-specific adaptive layers combined with a learned indicator function using straight-through estimator (STE) function~\cite{Yin2019UnderstandingSE}. TAPS demonstrated that selectively activating layers can preserve model performance while efficiently adapting it for multiple tasks, reducing catastrophic forgetting.

Building on TAPS, we propose an adaptive PEFT approach that selectively activates Low-Rank Adaptation (LoRA) blocks within models, particularly for vision and vision-language models like DINO~\cite{caron2021emerging} and CLIP~\cite{Radford2021LearningTV}. Our approach leverages an indicator function to dynamically gate LoRA blocks, achieving enhanced robustness while preserving zero-shot capabilities across multiple PEFT methods. Our experiments focus on LoRA~\cite{Hu2021LoRALA} and DoRA~\cite{Liu2024DoRAWL}, showcasing the flexibility of our method across different PEFT techniques.

Our results show that PEFT methods can preserve prior knowledge with fewer parameter changes than full fine-tuning, enhancing memory efficiency. However, catastrophic forgetting persists at higher ranks, which are often necessary for improved in-distribution accuracy. By selecting the most effective blocks, our approach achieves in-distribution accuracy comparable to full LoRA with as few as 6.25\% of active blocks, significantly improving computational efficiency. Figure~\ref{fig:intro} demonstrates that fully fine-tuning the CLIP model with FLYP~\cite{Goyal2022FinetuneLY} on ImageNet-1K, although enhancing in-distribution and domain-shifted performance, results in a notable drop in zero-shot retrieval (28\% decrease). LoRA trained at rank 128 does not improve these results and suffers from similar degradation. Our method, by selectively fine-tuning specific blocks, not only enhances in-distribution accuracy but also limits zero-shot performance drops to a maximum of 5.73\% compared to the pre-trained CLIP model. We conduct experiments across diverse settings to highlight various aspects of PEFT methods for fine-tuning vision and vision-language models, aiming to address related challenges through an indicator function.

Reducing the number of active LoRA blocks offers notable advantages, especially in applications like multi-task learning~\cite{Agiza2024MTLoRAAL,Feng2024MixtureofLoRAsAE} and image generation~\cite{Shah2023ZipLoRAAS,Zhong2024MultiLoRACF}, where directly merging LoRA blocks with the model’s original weights can compromise performance. Our method highlights the value of selectively activating fewer blocks, which not only preserves model integrity but also significantly reduces computational load. By minimizing the active block count, our approach achieves up to a 2.9x and 5x speed-up in inference times for LoRA and DoRA, respectively, while simultaneously lowering FLOPs. This efficiency gain is particularly beneficial in real-time and resource-constrained environments, enhancing scalability and responsiveness across diverse tasks.

Qualitatively, our LoRA-based adaptation is minimally parameter-intensive and largely model-agnostic. Quantitatively, our experiments reveal that this approach reduces catastrophic forgetting and maintains OOD performance across diverse tasks. By evaluating our method on both CLIP, a vision-language model, and ViT, a vision-specific model, we show that this PEFT enhancement not only builds on TAPS's strengths but also effectively addresses domain adaptation challenges, ensuring robustness and knowledge retention with broad applicability. We also highlight existing problems for future research.

\section{Related Works}
\paragraph{Parameter-Efficient Fine-Tuning (PEFT).} 
PEFT methods are categorized into Adapter-based, Prompt-based, and LoRA-variant approaches. Adapter-based methods (e.g., ~\cite{He2021TowardsAU, Davison2021CompacterEL, Houlsby2019ParameterEfficientTL}) introduce trainable modules into the frozen model backbone with minimal model changes. Prompt-based methods (e.g., ~\cite{Lester2021ThePO, Razdaibiedina2023ResidualPT, Jia2022VisualPT}) add soft tokens to the input but often struggle with initialization sensitivity, which affects their performance. Both Adapter- and Prompt-based methods tend to increase inference latency. LoRA-variants, such as AdaLoRA~\cite{Zhang2023AdaLoRAAB}, fine-tune the model with low-rank matrices, avoiding inference overhead. DoRA~\cite{Liu2024DoRAWL} enhances LoRA by decomposing weights into magnitude and direction for efficient updates, improving performance. MoRA~\cite{Jiang2024MoRAHU} uses high-rank square matrices for memory-intensive tasks, while VeRA~\cite{Kopiczko2023VeRAVR} reduces parameters by sharing low-rank matrices across layers. Tied-LoRA~\cite{Renduchintala2023TiedLoRAEP} ties matrices across layers to reduce complexity. We propose adding an indicator function to LoRA variants to boost efficiency and reduce catastrophic forgetting while enabling higher-rank training.

\vspace{-1mm}
\paragraph{Vision Foundation Models.}
Recent advances in vision-language models, such as ALIGN~\cite{Jia2021ScalingUV} and CLIP~\cite{Radford2021LearningTV}, have demonstrated impressive zero-shot transfer capabilities by leveraging contrastive learning objectives~\cite{Sohn2016ImprovedDM,Oord2018RepresentationLW} on large-scale datasets. Datasets like LAION-400M~\cite{Schuhmann2021LAION400MOD} and Conceptual Captions~\cite{Sharma2018ConceptualCA} have played a key role in pretraining these models by providing diverse image-text pairs, enabling generalization across various tasks and improving robustness to distribution shifts~\cite{Wortsman2021RobustFO,Fang2022DataDD}.  On the other hand, pure vision models, such as SimCLR~\cite{Chen2020ASF}, BYOL~\cite{Grill2020BootstrapYO}, and DINO~\cite{caron2021emerging}, have similarly leveraged self-supervised objectives to achieve strong feature representations~\cite{Baevski2022data2vecAG}, contributing to advancements in object recognition~\cite{yang2021instance,xie2021detco}, semantic segmentation~\cite{Pathak2016LearningFB,Liu2023GroundingDM}, super-resolution~\cite{wu2023practical}, and other tasks. Despite these strengths, both vision and vision-language models often encounter challenges when adapting to specialized domains, as large-scale pretraining datasets cannot fully capture all domain-specific nuances~\cite{Wang2024DoCA,Ni2023ContinualVR}. In this work, we explore PEFT techniques to robustly adapt vision models (e.g., DINO) and vision-language models (e.g., CLIP), aiming to mitigate catastrophic forgetting. This approach enhances adaptability across diverse domains and improves performance on domain-specific tasks.

\paragraph{Finetuning for Generalization.}
Maintaining generalization during finetuning of VLMs requires balancing in-distribution performance with robustness to domain shifts and zero-shot capabilities. For example, approaches for CLIP such as WiSE-FT~\cite{Wortsman2021RobustFO}, which combines original and finetuned model weights, LP-FT~\cite{Kumar2022FineTuningCD}, which uses a two-stage process to preserve pretrained features, and FLYP~\cite{Goyal2022FinetuneLY}, which mimics contrastive pretraining, focus mainly on distribution shifts. However, these methods often fail to fully address the significant degradation in OOD generalization after fine-tuning. Recently, ARF~\cite{Han2024AnchorbasedRF} introduced auxiliary supervision with rich semantic information to preserve the model's original feature space and enhance OOD generalization. Unlike ARF, our approach does not involve altering the data but instead leverages enhanced PEFT methods to address these challenges. Similarly, while SAFT~\cite{Nguyen2024SAFTTO} mitigates the loss of general knowledge by updating only a small subset of significant parameters in the CLIP model, our work extends PEFT techniques to various large vision-language foundation models, aiming for robust generalization across diverse tasks.
\section{Methodology}
\begin{figure}
    \centering
    \includegraphics[width=\linewidth]{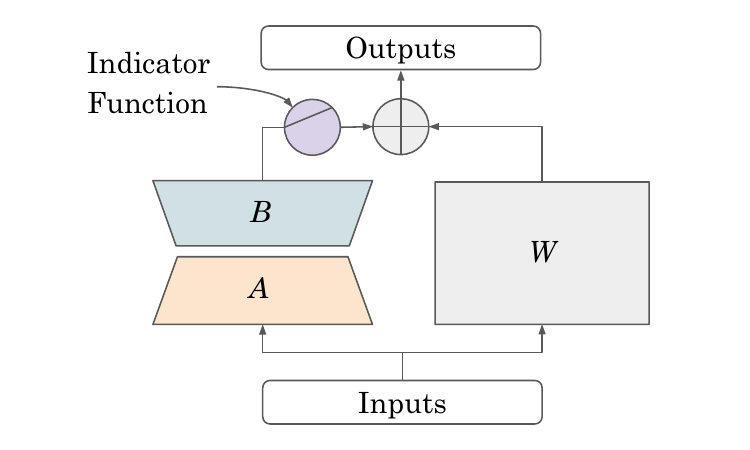}
    \caption{Overview of our proposed method incorporating LoRA with an indicator function. This approach allows flexible application of selective layer activation, making it compatible with various low-rank approximation techniques beyond LoRA.}
    \label{fig:intro_method}
\end{figure}

In the following subsections, we outline the problem setup and introduce our indicator function for LoRA-based methods, which selectively activates low-rank layers to reduce memory and computation, as depicted in Figure~\ref{fig:intro_method}.

\subsection{Problem Setup}

Following~\cite{AkbarianBafghi2024ParameterEF}, we begin with a self-supervised model pretrained on an unlabeled dataset \( D_{\text{pre}} = \{ x_{\text{pre}}^i \} \) sampled from a distribution \( P_{\text{pre}} \). The model is then fine-tuned on a labeled in-distribution dataset \( D_{\text{id}} = \{(x_{\text{id}}^i, y_{\text{id}}^i)\} \), drawn from distribution \( P_{\text{id}} \), where each sample \( x_{\text{id}}^i \) has an associated label \( y_{\text{id}}^i \in C_{\text{id}} \). After fine-tuning, the model is expected to perform well on a test dataset sampled from \( P_{\text{id}} \) with classes \( C_{\text{id}} \), as well as on the source dataset \( P_{\text{pre}} \). For evaluation on \( P_{\text{pre}} \), we apply k-nearest neighbors (K-NN) on the labeled test set of \( D_{\text{pre}} \).

For vision and language models, in line with~\cite{Han2024AnchorbasedRF}, we fine-tune a pretrained model on an in-distribution dataset \( D_{\text{id}} = \{(x_{\text{id}}^i, y_{\text{id}}^i)\} \), sampled from \( P_{\text{id}} \), where each image \( x_{\text{id}}^i \) is labeled as \( y_{\text{id}}^i \in C_{\text{id}} \). The fine-tuned model should perform comparably or better than traditional fine-tuning techniques on test data from \( P_{\text{id}} \) containing categories \( C_{\text{id}} \). We further assess the model’s OOD performance under both domain shift and zero-shot learning scenarios.  For domain shift, we use a dataset \( D_{\text{ds}} = \{(x_{\text{ds}}^i, y_{\text{ds}}^i)\} \), where images \( x_{\text{ds}}^i \) are drawn from a different distribution \( P_{\text{ds}} \) but share the same categories \( C_{\text{id}} \) as \( D_{\text{id}} \), so that \( P(X_{\text{id}}) \neq P(X_{\text{ds}}) \) while \( P(Y|X_{\text{id}}) = P(Y|X_{\text{ds}}) \). For zero-shot classification, we use \( D_{\text{zsc}} = \{(x_{\text{zsc}}^i, y_{\text{zsc}}^i)\} \), where images \( x_{\text{zsc}}^i \) belong to novel categories \( C_{\text{zsc}} \) not present in \( C_{\text{id}} \) (i.e., \( C_{\text{zsc}} \cap C_{\text{id}} = \emptyset \)). As an extension to~\cite{Han2024AnchorbasedRF}, we incorporate zero-shot retrieval, evaluating on a general dataset \( D_{\text{zsr}} = \{(x_{\text{zsr}}^i, t_{\text{zsr}}^i)\} \), where each image \( x_{\text{zsr}}^i \) is paired with a caption \( t_{\text{zsr}}^i \).

Our goal is to achieve high accuracy on \( D_{\text{id}} \) while preserving OOD generalization in both domain shift (\( D_{\text{ds}}\)) and zero-shot settings (\( D_{\text{zsr}}\) and \( D_{\text{zsc}}\)) for vision-language models, and on \( D_{\text{pre}} \) for vision models.

\subsection{Indicator Function for LoRA-Variant Methods}
LoRA~\cite{Hu2021LoRALA} is a technique for fine-tuning pre-trained models by learning a low-rank residual matrix \( \Delta W \), which is added to the fixed, original weights \( W_0 \). This allows the model to adapt to new tasks while maintaining a low memory footprint. The updated weights for each layer are computed as follows:

\begin{equation}
    W = W_0 + \Delta W
\end{equation}

where the residual \( \Delta W \) is factorized as \( \Delta W = AB \). Here, \( A \in \mathbb{R}^{m \times r} \) and \( B \in \mathbb{R}^{r \times n} \), with the rank \( r \ll \min(m, n) \) to ensure that memory usage remains minimal. This low-rank constraint enables the model to fine-tune with fewer parameters, which is particularly advantageous in memory-constrained environments.


Inspired by TAPS~\cite{Wallingford2022TaskAP}, which employs task-specific pruning by cloning the entire weight matrix and selectively activating task-relevant parameters, we decompose the weight adjustments into low-rank matrices \( A_\ell \) and \( B_\ell \), parameterized and controlled via a block-specific scoring parameter \( s_i \) for each block \( \ell = 1, \dots, L \). This scoring parameter facilitates the selective activation of LoRA blocks, tailoring the model's behavior to the relevance of each block for the specific task. The weight equation is modified as follows:

\begin{equation}
    W_\ell = W_{0,\ell} + I_\tau(s_i) A_\ell B_\ell,
\end{equation}

where \( W_{0,\ell} \) represents the pre-trained weights for block \( \ell \), and \( A_\ell B_\ell \) denotes the learned low-rank adjustment for that layer. The activation of each LoRA block is controlled by the STE indicator function \( I_\tau(s_i) \), defined as:

\begin{equation}
    I_\tau(s_i) =
    \begin{cases}
        1 & \text{if } s_i \geq \tau, \\
        0 & \text{otherwise}.
    \end{cases}
\end{equation}

Here, \( \tau \) is a threshold parameter that determines whether a block is significant for the task and should remain active. The scoring parameter \( s_i \) is initialized as part of the model's learnable parameters, enabling it to adapt dynamically during training to better suit the target task.

To encourage sparsity in the activation of LoRA blocks, we introduce a regularization term to the loss function, which penalizes non-zero values of \( s_i \). This term adds the \( \ell_1 \)-norm of \( s_i \) to the original loss function, which is formulated as follows:

\begin{equation}
    \mathcal{L}_{\text{total}} = \mathcal{L}_{\text{original}} + \lambda \sum_{i=1}^{L} |s_i|
\end{equation}

In this formulation, \( \mathcal{L}_{\text{original}} \) represents the task-specific loss (e.g., cross-entropy for classification), and \( \lambda \) controls the influence of regularization. Adjusting \( \lambda \) allows for fine-tuning the trade-off between task performance and memory efficiency by deactivating more LoRA blocks as \( \lambda \) increases. This reduces the memory and computational footprint while maintaining task-specific adaptability.

This formulation is general and can be applied to different LoRA methods, such as DoRA. By integrating an indicator function and sparsity-inducing regularization, this approach minimizes active LoRA blocks, balancing accuracy and efficiency. It provides a resource-friendly fine-tuning strategy, suitable for deployment in resource-constrained environments.

\section{Experiments}
\begin{figure*}
     \centering
     \includegraphics[width=\linewidth]{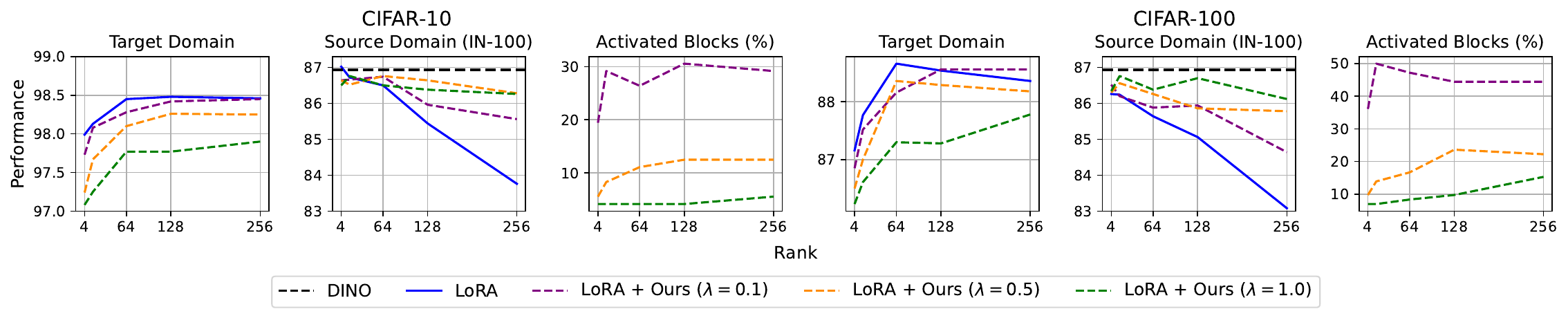}
     \caption{The figure compares the top-1 accuracy of pretrained and fine-tuned DINO ViT/S-16 models on CIFAR-10 and CIFAR-100 using LoRA and LoRA+Ours across various $\lambda$ values and different ranks. It demonstrates that our method achieves competitive performance on target datasets while preserving prior knowledge. Additionally, as $\lambda$ increases, the percentage of activated blocks decreases.}
     \label{fig:cifar}
\end{figure*}

In this section, we begin with experiments on vision models to examine the occurrence of catastrophic forgetting. Next, we scale up the experiments to include vision-and-language models, assessing both robustness and catastrophic forgetting. Additionally, we demonstrate the inference speed-up achieved with our method and visualize the most effective layers. Numerical results and hyperparameters for these experiments are provided in the Appendix.

\subsection{Vision Models}

\begin{figure*}
     \centering
     \includegraphics[width=\linewidth]{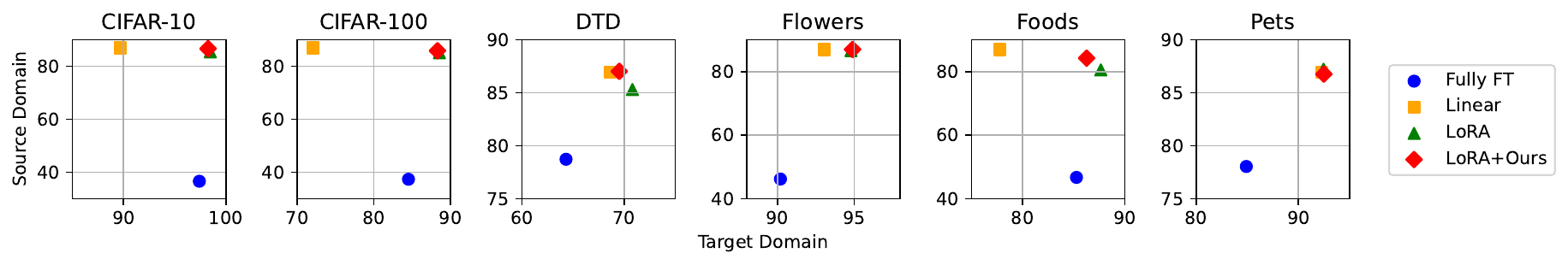}
     \caption{The figure compares top-1 accuracy of fine-tuned DINO ViT/S-16 models across transfer datasets and ImageNet-100. It demonstrates that models fine-tuned with LoRA, as well as those incorporating the indicator function, achieve high accuracy on target datasets (e.g., CIFAR-10) while retaining knowledge from the pre-trained dataset.}
     \label{fig:diffdatasets}
\end{figure*}
In this section, we conduct a simple experiment to assess the adaptability and resistance to catastrophic forgetting of a self-supervised vision model, specifically the pretrained DINO ViT/S-16 model~\cite{caron2021emerging}, originally trained on the ImageNet-1K (IN-1K) dataset. Fine-tuning is performed on five diverse target datasets: DTD~\cite{Cimpoi2013DescribingTI}, Flowers102~\cite{Nilsback2008AutomatedFC}, Food101~\cite{Bossard2014Food101M}, Pets37~\cite{Parkhi2012CatsAD}, CIFAR-10, and CIFAR-100~\cite{Krizhevsky2009LearningML}. Following the setup in~\cite{AkbarianBafghi2024ParameterEF}, we use K-NN to evaluate the fine-tuned model on the original pre-training data (source dataset), utilizing 100 classes of ImageNet-1K (IN-100) to improve efficiency and speed up evaluation. We set \( K = 20 \) for this assessment. The model's performance on both source and target datasets is compared to analyze catastrophic forgetting during new task learning. 

\begin{figure}
     \centering
     \includegraphics[width=\linewidth]{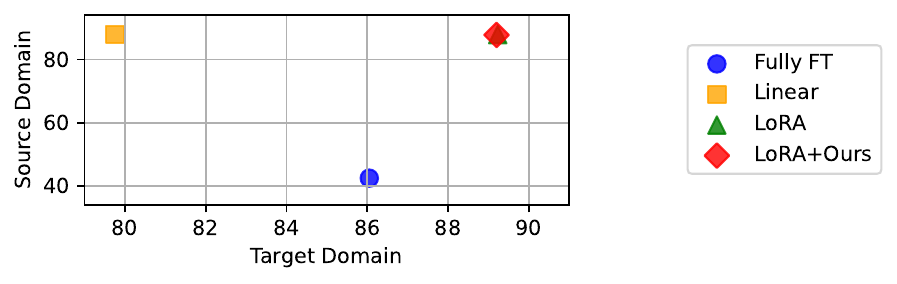}
     \caption{Comparison of top-1 accuracy for fine-tuned DINO ViT/B-16 models on CIFAR-100 and ImageNet-100. The results indicate that our method achieves similar performance regardless of model size.}
     \label{fig:cifar_size}
    \vspace{-0.2cm}
\end{figure}

We begin by conducting an experiment on CIFAR-10 and CIFAR-100 to evaluate the performance of our method across different ranks \{4, 16, 64, 128, 256\} and $\lambda$ values \{0.1, 0.5, 1.0\}. Figure~\ref{fig:cifar} demonstrates that LoRA shows a performance drop at higher ranks on both CIFAR-10 and CIFAR-100, with a decrease of approximately 3\% and 4\%, respectively, at rank 256. In contrast, our method achieves comparable accuracy in the target domain using fewer adaptors, while preserving knowledge on ImageNet. Specifically, at rank 256, our method with ($\lambda=0.1$) reduces forgetting to 0.98\% and 2.3\% on CIFAR-10 and CIFAR-100, respectively, using only 29.2\% and 44.4\% of blocks.

Next, we examine linear probing and full fine-tuning, alongside LoRA, on six target datasets. Both LoRA and our method, utilizing the indicator function, are trained with a rank of 128, and $\lambda$ is chosen from \{0.5, 1.0\}. Figure~\ref{fig:diffdatasets} demonstrates that our method achieves comparable results to LoRA across all datasets, highlighting its adaptability to various domains. Furthermore, our method achieves these results while activating only 12.5\%, 23.6\%, 2.7\%, 22.2\%, \( 5.5\%\), and \( 2.7\%\) of blocks for CIFAR-10, CIFAR-100, Flowers, Foods, Pets, and DTD, respectively.

To evaluate the efficiency of the selected blocks, we compared our method to a random selection of LoRA blocks. Our model, trained on CIFAR-100 at rank 16, activates 6.94\% (5 blocks) and 13.8\% (10 blocks) of blocks at the end of training with \( \lambda = 1 \) and \( \lambda = 0.5 \), respectively. We then randomly selected 5 and 10 blocks and trained the model with the same configuration as our method. Table~\ref{tab:random} shows that our method identifies more effective layers than random selection, achieving approximately 2\% higher accuracy on target dataset (CIFAR-100). However, as our method does not explicitly target blocks crucial for mitigating catastrophic forgetting, the performance on the source dataset (IN-100) remains comparable.

Finally, we evaluated our method on a larger version of DINO. Figure~\ref{fig:cifar_size} presents the results on CIFAR-100 using the DINO-ViT/B-16 model. Consistent with the findings from the ViT/S-16 model, our approach achieved accuracy comparable to LoRA while utilizing only 43\% of the LoRA blocks, demonstrating its scalability and efficiency on larger models.

\begin{table}
  \centering
  \begin{tabular}{@{}lcccc@{}}
    \toprule
    Method & Blocks (\%) & CIFAR-100 & IN-100 & Mean \\
    \midrule
    Ours & 6.94 & 86.61 & 86.76 & 86.68 \\
    Random & 6.94 & 84.52 & 86.50 & 85.51 \\
    \midrule
    Ours & 22.2 & 87.00 & 86.56 & 86.78\\
    Random & 22.2 & 85.46 & 86.04 & 85.75\\
    \bottomrule
  \end{tabular}
    \caption{Comparison of random selection versus our method for selecting activated blocks at rank 16. The results show that random selection fails to match the in-distribution accuracy achieved by our method.}
  \label{tab:random}
  \vspace{-3mm}
\end{table}

\subsection{Vision-Language Models}

\label{sec:vision-language}
\begin{figure*}
     \centering
     \includegraphics[width=\linewidth]{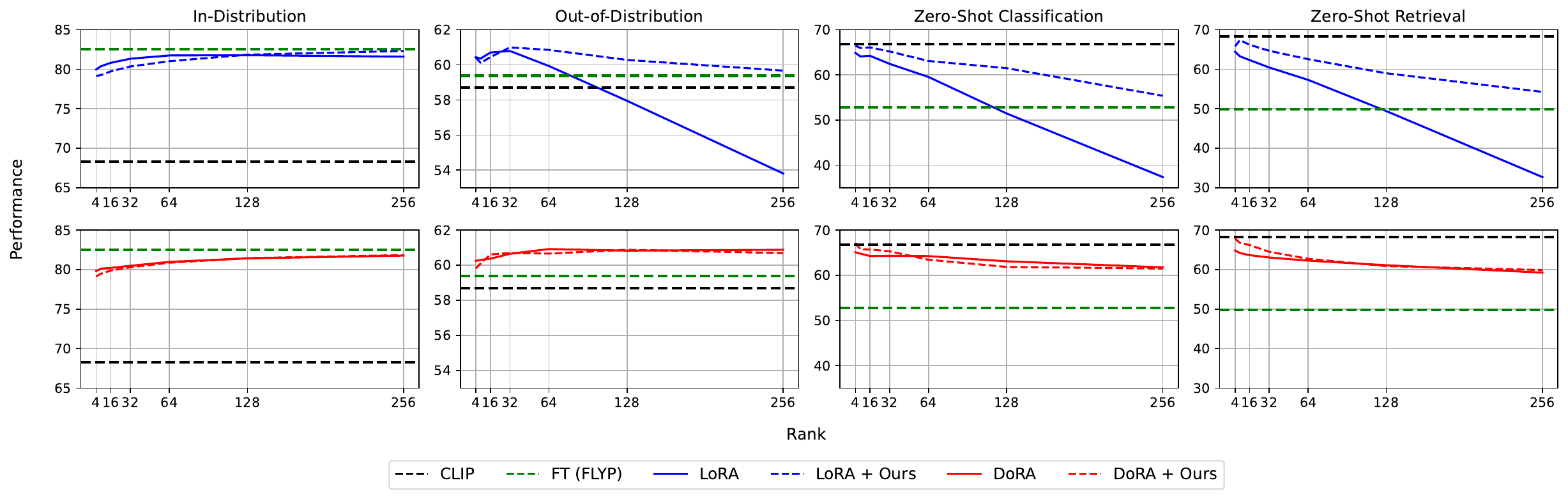}
     \caption{We compare in-distribution, out-of-distribution, zero-shot classification, and retrieval performance of the fine-tuned CLIP model with LoRA and DoRA at ranks 4, 16, 32, 64, 128, and 256, with/without the indicator function. All models were fine-tuned using FLYP. The FLYP-only model’s performance is shown as a green dashed line, and the pre-trained CLIP model as a grey dashed line. This comparison demonstrates that both PEFT methods can experience catastrophic forgetting at higher ranks, and the use of the indicator function can help mitigate this issue, particularly at those ranks. Rank 8 was omitted from the plot due to space constraints.}
     \label{fig:clip}
\end{figure*}

\begin{figure*}
     \centering
     \includegraphics[width=\linewidth]{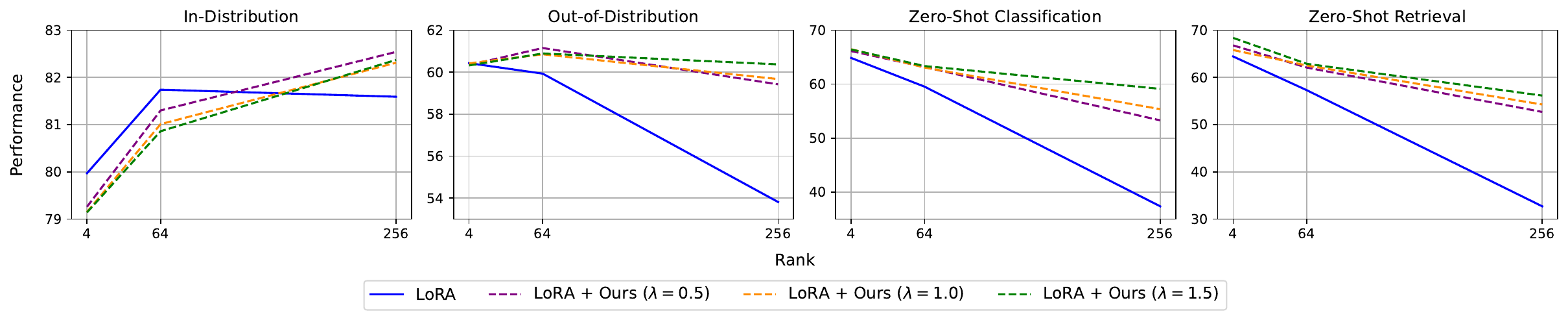}
     \vspace{-0.7cm}
     \caption{Comparison of in-distribution, out-of-distribution, zero-shot classification, and zero-shot retrieval performance of the fine-tuned CLIP model using LoRA at different ranks (4, 64, and 256), with and without the indicator function. All models were fine-tuned using FLYP. This comparison highlights that increasing $\lambda$ results in greater penalization, activating fewer LoRA blocks. This trade-off leads to reduced in-distribution accuracy but improved performance on other metrics.}
     \label{fig:lambda}
\end{figure*}

In this section, we scaled up the experiments to vision-language models and ImageNet-1K. We demonstrate the effectiveness of our method through an experiment designed to show not only the model’s performance on an in-distribution dataset but also its robustness to distribution shifts in both image and label spaces and catastrophic forgetting. CLIP is well-known for its zero-shot capability. We fine-tuned the CLIP model on ImageNet-1K~\cite{Deng2009ImageNetAL}, serving as our in-distribution dataset, and then evaluated it on an out-of-distribution benchmark to assess its robustness to image distribution shifts. Additionally, we conducted zero-shot classification and zero-shot retrieval experiments to examine the model’s ability to preserve prior knowledge. Our method was applied to both LoRA and DoRA, and we compared these results with those obtained by using LoRA and DoRA naively. Further details about our evaluation mechanism are provided in the following sections.
\paragraph{Out-of-Distribution Benchmark.} In this experiment, we use a diverse set of datasets, including ImageNet-A \cite{Hendrycks2019NaturalAE}, ImageNet-Sketch \cite{Wang2019LearningRG}, ObjectNet \cite{Barbu2019ObjectNetAL}, ImageNet-V2 \cite{Recht2019DoIC}, and ImageNet-R \cite{Hendrycks2020TheMF}, to evaluate our method when the categories remain the same but the distribution of images differs compared to ImageNet. For this benchmark, we report the average top-1 accuracy across the five robustness datasets.
\paragraph{Zero-Shot Classification Benchmark.} To evaluate the general capabilities of the CLIP model, we used a diverse set of commonly used transfer learning datasets, such as Aircraft \cite{Maji2013FineGrainedVC}, DTD~\cite{Cimpoi2013DescribingTI}, Flowers102 \cite{Nilsback2008AutomatedFC}, Food101~\cite{Bossard2014Food101M}, MNIST~\cite{Deng2012TheMD}, SUN97~\cite{Xiao2010SUNDL}, Pets37~\cite{Parkhi2012CatsAD}, STL10~\cite{Coates2011AnAO}, CIFAR-10~\cite{Krizhevsky2009LearningML}, and CIFAR-100 \cite{Krizhevsky2009LearningML}. These datasets have different distributions in both the image and label space. Template captions were used for classifying these images, and the reported number is the average accuracy across all datasets. 
\vspace{-4mm}
\paragraph{Zero-Shot Retrieval Benchmark.} We also utilized image-text retrieval (ITR) benchmarks to demonstrate the model's ability to preserve general knowledge, complementing our zero-shot classification benchmarks. The datasets include MS-COCO~\cite{Lin2014MicrosoftCC} and Flickr30k~\cite{Plummer2015Flickr30kEC}. We report the average performance of image-to-text and text-to-image retrieval for R@1 and R@5. R@K is a common ITR metric representing the proportion of correct matches in the top-K results.

\begin{figure}
     \centering
     \includegraphics[width=\linewidth]{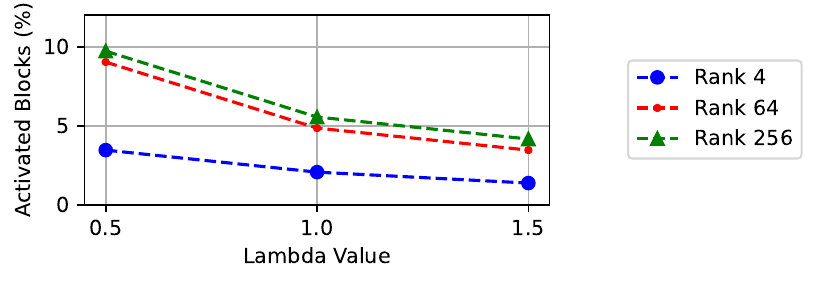}
     \vspace{-0.7cm}
     \caption{The y-axis represents the percentage of activated blocks after fine-tuning, while the x-axis indicates the $\lambda$ value. This figure demonstrates that as $\lambda$ increases, the number of activated blocks decreases. Additionally, models trained with a lower rank exhibit a smaller number of activated blocks.}
     \label{fig:clip_percentage}
     \vspace{-0.7cm}
\end{figure}
We fine-tuned the model using the hyperparameters specified in FLYP and applied the CLIP loss function. The value of $\lambda$ was set to 1 for both LoRA and DoRA, utilizing the indicator function. For training and zero-shot evaluation, we utilized the same text templates as those used in FLYP.  Figure~\ref{fig:clip} shows that catastrophic forgetting can occur even when using PEFT methods, particularly at higher ranks. Additionally, models tend to lose robustness to distribution shifts as the rank increases. This effect is especially pronounced when comparing LoRA and LoRA with the indicator function, where zero-shot classification and retrieval performance can drop by as much as 30\% and 40\%, respectively. However, using the indicator function helps the model retain its prior knowledge and even achieve higher in-distribution accuracy than when using LoRA alone. The percentage of activated blocks for LoRA using our indicator function ranges from 2.08\% to 6.25\% across different ranks. The percentage of activated blocks ranges from 1.39\% to 3.47\%, meaning that comparable results can be achieved with as few as one block. While higher ranks are generally used to achieve better in-distribution accuracy, our results highlight the trade-off: models may lose robustness and suffer from catastrophic forgetting.
\begin{figure}
     \centering
     \includegraphics[width=\linewidth]{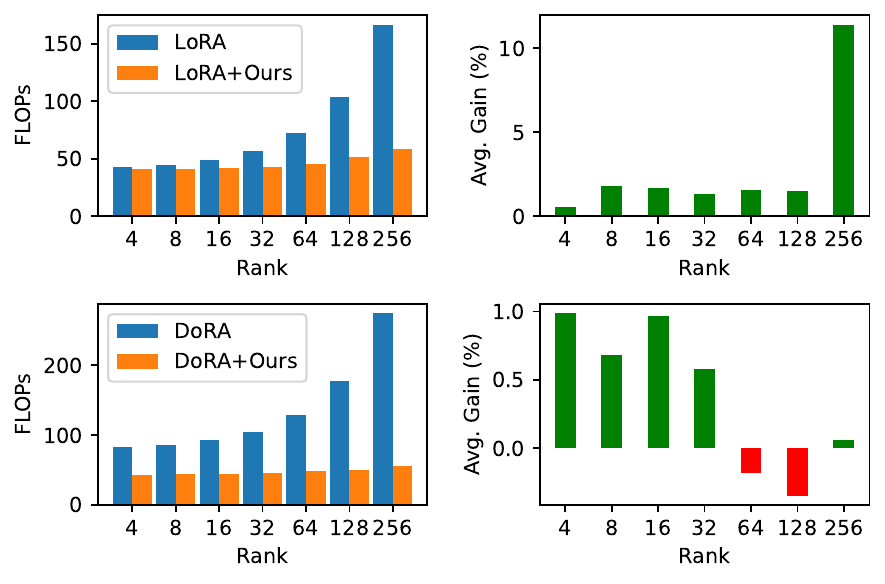}
     \vspace{-0.7cm}
    \caption{Comparison of FLOPs and Relative Gain for LoRA (first row) and DoRA (second row) with and without our method. Our method maintains constant FLOPs during inference, unlike other methods, where FLOPs increase with rank. LoRA generally shows higher relative gain than DoRA with our method, especially at higher ranks.}
     \label{fig:clip_flops}
     
\end{figure}

Next, we explore the effect of $\lambda$ on the number of activated blocks after fine-tuning the CLIP model. We replicate the experiment described in Section~\ref{sec:vision-language} using different values of $\lambda$. We select $\lambda$ from $\{0.5, 1, 1.5\}$ and rank from $\{4, 64, 256\}$ to investigate the behavior of LoRA and LoRA with the indicator function under various conditions.

Figure~\ref{fig:lambda} demonstrates that as \( \lambda \) increases, more blocks are deactivated, resulting in lower in-distribution accuracy but improved out-of-distribution and zero-shot performance. Conversely, decreasing \( \lambda \) enhances in-distribution accuracy at the cost of these other metrics. This highlights that adjusting \( \lambda \) provides a mechanism to balance model adaptation to the in-distribution dataset while mitigating catastrophic forgetting. Furthermore, Figure~\ref{fig:clip_percentage} shows that at higher ranks, more blocks are generally activated, whereas increasing \( \lambda \) reduces the percentage of activated blocks. These findings align with previous results observed in the DINO model.


\begin{figure}
     \centering
     \includegraphics[width=\linewidth]{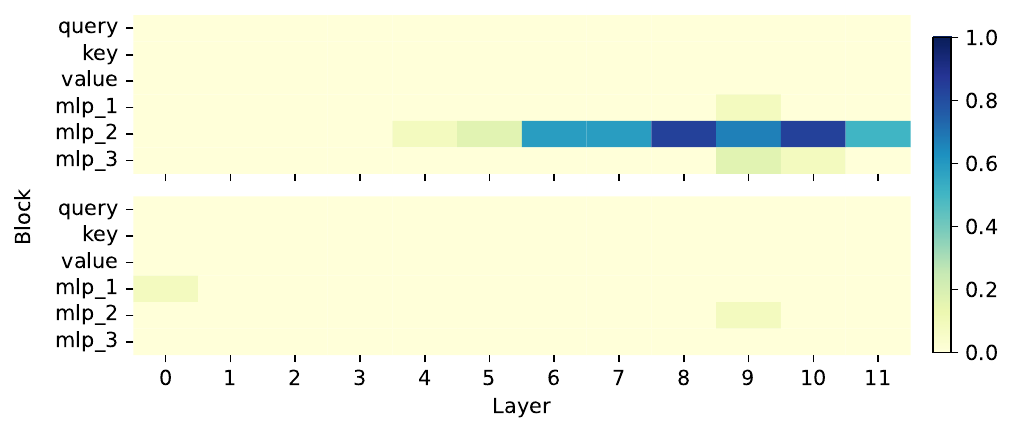}
     \vspace{-0.7cm}
    \caption{Visualization of the normalized count of activated blocks across 8 runs during fine-tuning of CLIP using our LoRA-based method. The top plot shows the vision transformer and the bottom the text transformer, with activations concentrated in the vision transformer’s higher layers, especially in the first MLP block.}
     \label{fig:clip_vis}
\end{figure}

\begin{figure}
     \centering
     \includegraphics[width=\linewidth]{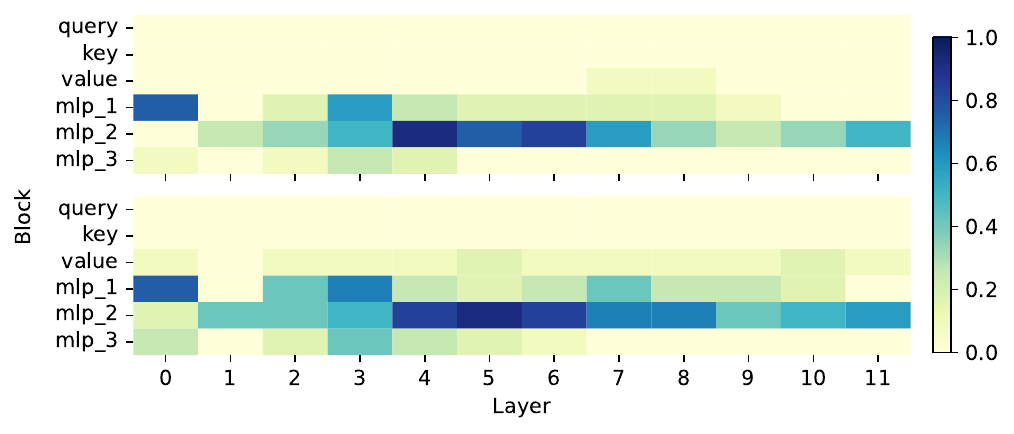}
     \vspace{-0.7cm}
    \caption{Normalized count of activated blocks across 15 runs for fine-tuning DINO-ViT/S with LoRA. The top plot shows fine-tuning on CIFAR-10 and the bottom on CIFAR-100, with similar trends: activations are primarily concentrated in the first MLP block of the feedforward layer.}
     \label{fig:vit_vis}
\end{figure}
\vspace{-0.7cm}
\subsection{Improvement in Inference Time}

In this subsection, we use floating-point operations per second (FLOPs) to quantify the inference computations of the CLIP model fine-tuned with LoRA and DoRA, both with and without the indicator function. By activating fewer blocks compared to the baselines, our method demonstrates improved inference efficiency, particularly when low-rank blocks are not merged with the base model.

To assess the relative performance improvement, we calculate the average in-distribution, out-of-distribution, and zero-shot accuracy of the models post-fine-tuning and present the differences. Figure~\ref{fig:clip_flops} displays the FLOPs and performance gains across varying ranks. Unlike traditional LoRA and DoRA approaches, which require significantly higher FLOPs at larger ranks during inference due to unmerged low-rank weights, our method sustains a relatively constant FLOP count. For instance, at rank 256, our approach achieves approximately 2.9x and 5x faster inference compared to LoRA and DoRA, respectively, without sacrificing performance. Additionally, our method yields a higher relative gain for LoRA than for DoRA.

\subsection{The Most Effective Layers}

In this section, we analyze which components of the transformer architecture in the CLIP and DINO-ViT/S models are most frequently activated during fine-tuning. For CLIP, we report the activation of blocks over 7 runs at ranks 4, 8, 16, 32, 64, 128, and 256 with $\lambda = 1$, along with an additional 6 runs at ranks 4, 64, and 256 with $\lambda$ values of 0.5 and 1.5. For DINO-ViT/S, we report activation counts over 15 runs, fine-tuned separately on CIFAR-10 and CIFAR-100, using ranks 4, 16, 64, 128, and 256 with $\lambda$ values of 0.1, 0.5, and 1. All activation counts are normalized by the number of runs. In each experiment, we add 6 LoRA blocks per layer: one each for the query, key, value, and attention MLP (mlp\_1), and two for the MLPs in the feedforward layer (mlp\_2 and mlp\_3).

Figure~\ref{fig:clip_vis} presents the visualization of activated blocks in the fine-tuned CLIP model. The figure indicates that most activated blocks are located within the vision transformer, suggesting that the model does not rely heavily on the text transformer to adapt to new datasets. Additionally, a majority of activations occur in the mlp\_2 blocks. Figure~\ref{fig:vit_vis} shows the activated blocks for the DINO-ViT/S model, where we observe that the activation pattern in mlp\_2 aligns with that of the CLIP model. However, in DINO-ViT/S, activations are distributed across layers, whereas in CLIP, activations are more concentrated in the final layers. In both examples, the query, key, and value components appear to be less frequently selected, possibly because they contain fewer parameters than the feedforward layers in transformers. Consequently, the method prioritizes activating the feedforward layers, allowing for a greater degree of parameter modification.

\subsection{Different Losses for Regularization}

\begin{table}
  \centering
  \begin{tabular}{@{}lcccc@{}}
    \toprule
    Loss & Blocks (\%) & CIFAR-100 & IN-100 & Mean \\
    \midrule
    \( \ell_1 \)-norm & 9.7 & 87.28 & 86.70 & 86.99 \\
    \( \ell_2 \)-norm & 11.1 & 87.47 & 86.32 & 86.89 \\
    Hinge & 22.2 & 88.03 & 86.26 & 87.14\\
    \bottomrule
  \end{tabular}
  \caption{Comparison of regularization losses for reducing the percentage of activated blocks in our method. Each regularization approach effectively decreases the number of activated blocks while maintaining competitive accuracy on CIFAR-100 and IN-100.}
  \vspace{-0.3cm}
  \label{tab:diffloss}
\end{table}

We explore the effect of different regularization losses on optimizing indicator values, specifically comparing \( \ell_1 \)-norm, \( \ell_2 \)-norm, and hinge loss. The \( \ell_1 \)-norm promotes sparsity through \( \lambda \sum_{i=1}^{L} |s_i| \), encouraging minimal block activation. The \( \ell_2 \)-norm, expressed as \( \lambda \sum_{i=1}^{L} s_i^2 \), encourages smoothness and more gradual value distributions. Hinge loss, defined as \( \lambda \sum_{i=1}^{L} \max(0, s_i - \tau) \), restricts values within a target range without penalizing those below the threshold. We trained the DINO model on CIFAR-100 with each of these loss functions, applying LoRA at rank 128 both with and without our method.

Table~\ref{tab:diffloss} compares the performance of these approaches. Each regularization loss effectively reduces the number of activated blocks, allowing us to select a suitable loss for further experimentation. Notably, hinge loss tends to retain a higher number of blocks since it does not penalize values below the threshold. A limitation of our method is the lack of precise control over the final number of activated blocks prior to training; instead, this is adjusted indirectly by tuning the  \( \lambda \) values.

\section{Conclusion}
In this work, we demonstrated that PEFT methods are highly effective in mitigating catastrophic forgetting. However, they can still exhibit forgetting, particularly at higher ranks, which are critical for achieving high in-distribution accuracy. To address this, we introduced an extension to TAPS for PEFT, applying an indicator function to selectively activate LoRA blocks. This simple approach can be applied to various PEFT methods (e.g., DoRA), enabling the adaptation of pre-trained models to new domains while maintaining robustness to distribution shifts. Our method preserves generalization in OOD and minimizes the knowledge loss typical of traditional fine-tuning methods.

With effective performance using as few as 5\% of active blocks, our method significantly reduces computational costs during inference, demonstrating both efficiency and adaptability. We fine-tuned and evaluated DINO on six diverse datasets to demonstrate the adaptability of our approach. Additionally, we fine-tuned the CLIP model on ImageNet-1K and evaluated it on 17 datasets across different tasks (five for domain shift, ten for zero-shot classification, and two for zero-shot retrieval). The results indicate that our method achieves competitive performance compared to other PEFT methods on in-distribution tasks, while preserving prior knowledge and maintaining robustness to distribution shifts.

While our method proves broadly effective on models such as CLIP and DINO, its primary objective is not explicitly designed to mitigate catastrophic forgetting. Although it minimizes parameter changes by activating the most effective blocks, it does not directly address catastrophic forgetting. Furthermore, our method lacks control over the exact number of activated blocks, which presents an additional limitation. Future research could explore optimized block activation strategies and applications across other pre-trained architectures. Despite these limitations, we believe this work contributes to a deeper understanding of PEFT methods for domain adaptation in the vision field and presents a promising, efficient approach for adapting pre-trained models to new domains.

{
    \small
    \bibliographystyle{ieeenat_fullname}
    \bibliography{main}
}

\newpage
\appendix
\clearpage
\setcounter{page}{1}
\maketitlesupplementary

\section{Overview}
\label{sec:rationale}

This supplementary document complements the main paper by:

\begin{enumerate}
\item Providing the hyperparameters used in experiments for vision and vision-language models.
\item Presenting the numerical results corresponding to the figures included in the main paper.
\end{enumerate}

\section{Hyperparameters} For experiments with DINO, the model was trained for 5000 steps, with evaluations every 100 steps. The best checkpoint, determined by validation set performance, was used for evaluation. We used a learning rate of 0.005 with the SGD optimizer and a cosine learning rate scheduler, including a 500-step warmup phase. The same hyperparameters were applied across all CLIP experiments, using the AdamW optimizer with a learning rate of 0.00001, a weight decay of 0.1, and a cosine scheduler with 500 warmup steps. All models were trained for 10 epochs, and the final checkpoint was used for evaluation. We froze the layers of the text and vision transformers, unfreezing only the remaining parameters for both LoRA and LoRA combined with our method.

\section{Numerical Results} This section presents numerical results to complement the figures in the main paper, providing detailed accuracy scores for different datasets and experimental conditions.

Table~\ref{tab:c10-appendix} provides the accuracy results for models fine-tuned on CIFAR-10 and evaluated on ImageNet-100 (IN-100). The table includes results across various ranks and different values of $\lambda$, which controls the trade-off between the task performance and the amount of parameter adaptation. The table also highlights the improvements achieved when LoRA is combined with our proposed method compared to standalone Linear Probing and Full Fine-Tuning (FT). Table~\ref{tab:c100-appendix} extends this analysis to CIFAR-100, evaluated on ImageNet-100 (IN-100), highlighting the performance of LoRA combined with our method at different ranks and $\lambda$ values. Table~\ref{tab:biggermodel} presents results for DINO with a larger backbone (ViT-B/16), where models are fine-tuned on CIFAR-100 and evaluated on ImageNet-100. 

Tables~\ref{tab:dtd-appendix},\ref{tab:pets-appendix},\ref{tab:flowers-appendix}, and~\ref{tab:foods-appendix} show the accuracy on target datasets (DTD, Pets37, Flowers102, and Foods101) and the corresponding source dataset (IN-100). These tables provide evidence that our approach maintains performance on source datasets while improving generalization on target datasets.

Tables~\ref{tab:lora-ds} and~\ref{tab:dora-ds} present the performance of models fine-tuned on ImageNet-1K (IN-1K) and evaluated on distribution-shift datasets, including ImageNet-A (IN-A), ImageNet-Sketch (IN-S), ImageNet-R (IN-R), ImageNet-V2, and ObjectNet (ON).

Tables~\ref{tab:lora-cls} and~\ref{tab:dora-cls} provide a detailed comparison of the models' performance on IN-1K and zero-shot classification datasets, including CIFAR-10 (C10), CIFAR-100 (C100), DTD, MNIST (MNI), SUN, Aircrafts (AIR), Flowers (FLW), Foods (FOD), and Pets.

Finally, Tables~\ref{tab:lora-ret} and~\ref{tab:dora-ret} present numerical results for performance on IN-1K and zero-shot retrieval datasets, including MS-COCO and Flickr-30K for image-to-text retrieval (IR@1 and IR@5) and text-to-image retrieval (TR@1 and TR@5).

\begin{table*}
  \centering
  \begin{tabular*}{\linewidth}{@{\extracolsep{\fill}} lcccccc}
    \toprule
    Method & R & $\lambda$ & Blocks (\%) & CIFAR-10 & IN-100 & Mean \\
    \midrule
    Linear & - & - & - & 89.71 & 87.20  & 88.45 \\
    \midrule
    FT & - & - & - & 97.41 & 36.58 & 66.99 \\
    \midrule
    \multirow{5}{*}{LoRA} & 4&-&100&97.99&87.02&92.50\\
&16&-&100&98.13&86.72&92.42\\
&64&-&100&98.45&86.5&92.47\\
&128&-&100&98.48&85.44&91.96\\
&256&-&100&98.46&83.76&91.11\\
    \midrule
    \multirow{15}{*}{+Ours} 
&4&1&4.16&97.08&86.50&91.79\\
&16&1&4.17&97.25&86.76&92.01\\
&64&1&4.17&97.77&86.50&92.14\\
&128&1&4.17&97.77&86.38&92.08\\
&256&1&5.56&97.90&86.26&92.08\\
&4&0.5&5.55&97.24&86.64&91.94\\
&16&0.5&8.30&97.67&86.52&92.10\\
&64&0.5&11.11&98.10&86.76&92.43\\
&128&0.5&12.50&98.26&86.64&92.45\\
&256&0.5&12.50&98.25&86.28&92.27\\
&4&0.1&19.44&97.73&86.64&92.19\\
&16&0.1&29.17&98.08&86.66&92.37\\
&64&0.1&26.39&98.28&86.74&92.51\\
&128&0.1&30.56&98.42&85.96&92.19\\
&256&0.1&29.17&98.45&85.56&92.01\\
    \bottomrule
  \end{tabular*}
    \caption{Comparison of the performance of Linear Probing, Full Fine-Tuning (FT), LoRA, and LoRA combined with our method across various ranks (4, 16, 64, 128, and 256) and lambda values (0.1, 0.5, and 1). Models are fine-tuned on CIFAR-10 and evaluated on ImageNet-100 (IN-100). \textit{R} denotes the rank, and \textit{Blocks (\%)} indicates the percentage of activated blocks.}
  \label{tab:c10-appendix}
\end{table*}

\begin{table*}
  \centering
  \begin{tabular*}{\linewidth}{@{\extracolsep{\fill}} lcccccc}
    \toprule
    Method & R & $\lambda$ & Blocks (\%) & CIFAR-100 & IN-100 & Mean \\
    \midrule
    Linear & - & - & - & 72.07 & 87.20 & 88.32 \\
    \midrule
    FT & - & - & - & 84.51 & 37.34 & 60.92 \\
    \midrule
    \multirow{5}{*}{LoRA} &4&-&100&87.16&86.26&86.71\\
&16&-&100&87.77&86.24&87.01\\
&64&-&100&88.66&85.64&87.15\\
&128&-&100&88.54&85.06&86.80\\
&256&-&100&88.36&83.08&85.72\\
    \midrule
    \multirow{15}{*}{+Ours} 
&4&1&6.94&86.23&86.34&86.29\\
&16&1&6.94&86.61&86.76&86.69\\
&64&1&8.33&87.3&86.38&86.84\\
&128&1&9.72&87.28&86.70&86.99\\
&256&1&15.28&87.78&86.12&86.95\\
&4&0.5&9.72&86.50&86.30&86.40\\
&16&0.5&13.89&87.00&86.56&86.78\\
&64&0.5&16.66&88.36&86.26&87.31\\
&128&0.5&23.61&88.29&85.86&87.08\\
&256&0.5&22.22&88.18&85.78&86.98\\
&4&0.1&36.11&86.85&86.52&86.69\\
&16&0.1&50.00&87.52&86.20&86.86\\
&64&0.1&47.22&88.16&85.88&87.02\\
&128&0.1&44.44&88.56&85.94&87.25\\
&256&0.1&44.44&88.56&84.64&86.60\\
    \bottomrule
  \end{tabular*}
    \caption{Comparison of the performance of Linear Probing, Full Fine-Tuning (FT), LoRA, and LoRA combined with our method across various ranks (4, 16, 64, 128, and 256) and lambda values (0.1, 0.5, and 1). Models are fine-tuned on CIFAR-100 and evaluated on ImageNet-100 (IN-100). \textit{R} denotes the rank, and \textit{Blocks (\%)} indicates the percentage of activated blocks.}
  \label{tab:c100-appendix}
\end{table*}

\begin{table*}
  \centering
  \begin{tabular*}{\linewidth}{@{\extracolsep{\fill}} lcccccc}
    \toprule
    Method & R & $\lambda$ & Blocks (\%) & CIFAR-100 & IN-100 & Mean \\
    \midrule
    Linear & - & - & - & 79.75 & 87.84 & 83.79 \\
    \midrule
    FT & - & - & - & 86.05 & 42.50 & 64.27 \\
    \midrule
    LoRA &128&-&100&89.23&87.68&88.46\\
    \midrule
    \multirow{3}{*}{+Ours} 
    &128&1&6.94&88.34&87.96&88.15\\
&128&0.5&9.72&88.33&87.80&88.07\\
&128&0.1&43.06&89.20&87.70&88.45\\
    \bottomrule
  \end{tabular*}
    \caption{Performance comparison of Linear Probing, Full Fine-Tuning (FT), LoRA, and LoRA combined with our method at rank 128 and lambda values (0.1, 0.5, and 1). The models are fine-tuned on CIFAR-100 and evaluated on ImageNet-100 (IN-100). \textit{R} denotes the rank, and \textit{Blocks (\%)} indicates the percentage of activated blocks. All models use ViT-B/16 as the backbone.}
  \label{tab:biggermodel}
\end{table*}

\begin{table*}
  \centering
  \begin{tabular*}{\linewidth}{@{\extracolsep{\fill}} lcccccc}
    \toprule
    Method & R & $\lambda$ & Blocks (\%) & DTD & IN-100 & Mean \\
    \toprule
    Linear & - & - & - & 68.62 & 87.20 & 77.91 \\
    \midrule
    FT & - & - & - & 64.31 & 78.72 & 71.51 \\
    \midrule
    LoRA &128&-&100&70.8&85.30&78.05\\
    \midrule
    +Ours
    &128&1&2.70&69.52&87.02&78.27\\
    \bottomrule
  \end{tabular*}
    \caption{Performance comparison of Linear Probing, Full Fine-Tuning (FT), LoRA, and LoRA combined with our method at rank 128 and a lambda value of 1. The models are fine-tuned on DTD and evaluated on ImageNet-100 (IN-100). \textit{R} denotes the rank, and \textit{Blocks (\%)} indicates the percentage of activated blocks.}
  \label{tab:dtd-appendix}
\end{table*}

\begin{table*}
  \centering
  \begin{tabular*}{\linewidth}{@{\extracolsep{\fill}} lcccccc}
    \toprule
    Method & R & $\lambda$ & Blocks (\%) & Pets37 & IN-100 & Mean \\
    \toprule
    Linear & - & - & - & 92.26 & 87.20 & 89.73 \\
    \midrule
    FT & - & - & - &  84.90 & 78.04 & 81.47 \\
    \midrule
    LoRA &128&-&100&92.45&87.22&89.83\\
    \midrule
    +Ours
    &128&0.5&5.55&92.48&86.76&89.62\\
    \bottomrule
  \end{tabular*}
    \caption{Performance comparison of Linear Probing, Full Fine-Tuning (FT), LoRA, and LoRA combined with our method at rank 128 and a lambda value of 0.5. The models are fine-tuned on Pets37 and evaluated on ImageNet-100 (IN-100). \textit{R} denotes the rank, and \textit{Blocks (\%)} indicates the percentage of activated blocks.}
  \label{tab:pets-appendix}
\end{table*}

\begin{table*}
  \centering
  \begin{tabular*}{\linewidth}{@{\extracolsep{\fill}} lcccccc}
    \toprule
    Method & R & $\lambda$ & Blocks (\%) & Flowers & IN-100 & Mean \\
    \toprule
    Linear & - & - & - & 93.04 & 87.20 & 90.12 \\
    \midrule
    FT & - & - & - &  90.20 & 46.18 & 68.19 \\
    \midrule
    LoRA &128&-&100&94.80&86.58&90.69\\
    \midrule
    +Ours
    &128&0.5&2.77&94.51&86.92&90.71\\
    \bottomrule
  \end{tabular*}
    \caption{Performance comparison of Linear Probing, Full Fine-Tuning (FT), LoRA, and LoRA combined with our method at rank 128 and a lambda value of 0.5. The models are fine-tuned on Flowers102 and evaluated on ImageNet-100 (IN-100). \textit{R} denotes the rank, and \textit{Blocks (\%)} indicates the percentage of activated blocks.}
  \label{tab:flowers-appendix}
\end{table*}

\begin{table*}
  \centering
  \begin{tabular*}{\linewidth}{@{\extracolsep{\fill}} lcccccc}
    \toprule
    Method & R & $\lambda$ & Blocks (\%) & Foods & IN-100 & Mean \\
    \toprule
    Linear & - & - & - & 77.80 & 87.20 & 82.50 \\
    \midrule
    FT & - & - & - &  85.27 & 46.70 & 65.98 \\
    \midrule
    LoRA &128&-&100&87.65&80.50&84.07\\
    \midrule
    +Ours
    &128&0.5&22.77&86.76&84.08&85.42\\
    \bottomrule
  \end{tabular*}
    \caption{Performance comparison of Linear Probing, Full Fine-Tuning (FT), LoRA, and LoRA combined with our method at rank 128 and a lambda value of 0.5. The models are fine-tuned on Foods101 and evaluated on ImageNet-100 (IN-100). \textit{R} denotes the rank, and \textit{Blocks (\%)} indicates the percentage of activated blocks.}
  \label{tab:foods-appendix}
\end{table*}

\begin{table*}
\begin{center}
\begin{tabular*}{\linewidth}{@{\extracolsep{\fill}} lcccccccccc}
\toprule
\multirow{2}{*}{Model} & \multirow{2}{*}{R} & \multirow{2}{*}{Blocks (\%)} & \multirow{2}{*}{IN-1K} & \multicolumn{6}{c}{Distribution Shifts Datasets} & \multirow{2}{*}{Mean} \\
\cmidrule{5-10}
 & & & & IN-A & IN-V2 & IN-S & IN-R & ON & Mean & \\
\toprule
CLIP & - & - & 68.30&50.00&61.90&48.30&77.70&55.40&58.70&63.50\\
\midrule
FLYP & - & - &82.53&48.29&73.25&49.38&71.34&54.63&59.4&70.95\\
\midrule
\multirow{7}{*}{LoRA} & 4&100&79.97&51.44&71.18&48.93&75.06&55.56&60.43&70.20\\
&8&100&80.38&51.35&71.34&49.00&74.60&55.51&60.36&70.37\\
&16&100&80.81&51.47&71.89&49.52&74.75&55.89&60.70&70.76\\
&32&100&81.33&52.00&72.27&49.75&73.81&56.10&60.79&71.06\\
&64&100&81.74&50.00&72.52&49.45&72.45&55.21&59.93&70.83\\
&128&100&81.77&45.97&72.53&48.23&69.57&53.47&57.95&69.86\\
&256&100&81.59&38.55&71.37&44.81&64.01&50.30&53.81&67.70\\

\midrule
\multirow{7}{*}{+Ours} &4&2.08&79.15&51.07&70.16&49.13&76.08&55.67&60.42&69.79\\
&8&2.08&79.26&51.12&70.22&48.45&75.49&55.29&60.11&69.69\\
&16&2.78&79.73&51.44&70.79&48.75&75.70&55.51&60.44&70.08\\
&32&3.47&80.35&51.72&71.67&49.94&75.46&56.16&60.99&70.67\\
&64&4.86&81.01&51.17&72.05&50.15&75.03&55.83&60.85&70.93\\
&128&6.25&81.84&50.35&72.42&49.84&73.42&55.36&60.28&71.06\\
&256&5.56&82.31&48.89&72.93&49.23&72.18&55.11&59.67&70.99\\

\bottomrule
\end{tabular*}
\end{center}
\vspace{-0.5cm}
\caption{Comparison of the performance of CLIP, Full Fine-Tuning (FLYP), LoRA, and LoRA combined with our method across different ranks (4, 16, 32, 64, 128, and 256) with a lambda value of 1. The models are fine-tuned on ImageNet-1K and evaluated on distribution datasets. \textit{R} denotes the rank, and \textit{Blocks (\%)} indicates the percentage of activated blocks.}
\label{tab:lora-ds}
\end{table*}

\begin{table*}
\begin{center}
\begin{tabular*}{\linewidth}{@{\extracolsep{\fill}} lcccccccccc}
\toprule
\multirow{2}{*}{Model} & \multirow{2}{*}{R} & \multirow{2}{*}{Blocks (\%)} & \multirow{2}{*}{IN-1K} & \multicolumn{6}{c}{Distribution Shifts Datasets} & \multirow{2}{*}{Mean} \\
\cmidrule{5-10}
 & & & & IN-A & IN-V2 & IN-S & IN-R & ON & Mean & \\
\toprule
CLIP & - & - & 68.30&50.00&61.90&48.30&77.70&55.40&58.70&63.50\\
\midrule
FLYP & - & - &82.53&48.29&73.25&49.38&71.34&54.63&59.4&70.95\\
\midrule
\multirow{7}{*}{DoRA} &4&100&79.81&51.00&70.77&49.01&75.34&55.14&60.25& 
70.03\\
&8&100&80.12&51.37&70.90&48.91&75.10&55.22&60.30& 70.21\\
&16&100&80.22&51.43&71.21&49.02&74.74&55.47&60.37&70.30\\
&32&100&80.48&52.01&71.55&49.38&74.78&55.55&60.65&70.57\\
&64&100&80.98&52.69&71.79&49.60&74.75&55.76&60.92&70.95\\
&128&100&81.42&51.89&72.34&49.71&74.32&55.91&60.83&71.13\\
&256&100&81.77&52.12&72.39&49.96&73.77&56.15&60.88&71.32\\

\midrule
\multirow{7}{*}{+Ours} &4&1.39&79.13&50.55&70.23&48.36&75.22&54.74&59.82&69.48\\
&8&2.08&79.49&50.91&70.57&48.66&75.30&55.07&60.10&69.80\\
&16&2.08&79.88&51.73&71.17&49.00&75.37&55.79&60.61&70.25\\
&32&2.78&80.29&51.01&71.35&49.74&75.29&56.12&60.70&70.50\\
&64&3.47&80.87&50.99&71.86&49.86&74.61&55.98&60.66&70.77\\
&128&3.47&81.45&51.53&72.10&50.39&74.26&56.13&60.88&71.17\\
&256&3.47&81.84&50.80&72.71&50.35&73.69&55.89&60.69&71.26\\

\bottomrule
\end{tabular*}
\end{center}
\vspace{-0.5cm}
\caption{Comparison of the performance of CLIP, Full Fine-Tuning (FLYP), DoRA, and DoRA combined with our method across different ranks (4, 16, 32, 64, 128, and 256) with a lambda value of 1. The models are fine-tuned on ImageNet-1K and evaluated on distribution datasets. \textit{R} denotes the rank, and \textit{Blocks (\%)} indicates the percentage of activated blocks.}
\label{tab:dora-ds}
\end{table*}

\begin{table*}
\begin{center}
\begin{tabular*}{\linewidth}{@{\extracolsep{\fill}} lccccccccccccccccccccccc}
\toprule
\multirow{2}{*}{Model} & \multirow{2}{*}{R} & \multirow{2}{*}{Blocks (\%)} & \multirow{2}{*}{IN-1K} & \multicolumn{10}{c}{Zero-Shot Classification Datasets} & \multirow{2}{*}{Mean} \\
\cmidrule{5-14}
 & & & & C10 & C100 & DTD & MNI & SUN & AIR & FLW & FOD & Pets & Mean & \\
\toprule
CLIP & - & - &  68.3&90.8&68.3&44.7&59.4&65.3&24.3&71.0&88.5&89.1&66.8&67.6\\
\midrule
FLYP & - & - &  82.5&89.7&64.9&35.6&44.6&52.8&8.5&39.1&62.7&77.3&52.8&67.7\\
\midrule
\multirow{7}{*}{LoRA} 
&4&100&80.0&92.4&69.1&44.1&54.7&65.1&19.6&65.6&84.3&88.9&64.9&72.4\\
&8&100&80.4&92.2&68.0&43.5&51.8&64.4&19.1&65.7&83.6&88.6&64.1&72.2\\
&16&100&80.8&91.9&68.4&43.5&55.8&63.2&18.8&65.5&82.5&88.3&64.2&72.5\\
&32&100&81.3&91.9&68.0&42.9&51.9&61.3&16.5&61.1&79.8&88.7&62.4&71.9\\
&64&100&81.7&90.6&66.2&41.9&45.3&58.9&15.7&54.1&75.8&87.4&59.5&70.6\\
&128&100&81.8&88.4&62.4&35.5&43.7&51.4&7.7&35.6&61.0&77.4&51.4&66.6\\
&256&100&81.6&85.8&57.2&22.9&18.8&36.7&2.1&15.3&29.4&68.1&37.4&59.5\\
\midrule
\multirow{7}{*}{+Ours} 
&4&2.08&79.2&92.5&69.6&45.0&62.2&65.3&21.3&67.9&85.7&88.7&66.5&72.8\\
&8&2.08&79.3&92.5&69.5&46.2&56.2&65.3&22.1&67.3&85.9&88.6&65.9&72.6\\
&16&2.78&79.7&92.3&69.6&45.9&62.1&64.3&20.8&66.6&84.4&88.8&66.1&72.9\\
&32&3.47&80.4&92.1&68.9&46.1&59.1&63.3&19.9&65.6&83.1&88.6&65.2&72.8\\
&64&4.86&81.0&91.4&68.2&46.0&50.5&61.7&19.1&62.6&79.9&88.3&63.1&72.0\\
&128&6.25&81.8&91.6&67.1&43.5&52.7&59.6&17.6&58.0&75.9&87.4&61.5&71.7\\
&256&5.56&82.3&90.6&66.4&39.6&46.3&54.5&11.0&42.3&66.0&81.7&55.4&68.8\\
\bottomrule
\end{tabular*}
\end{center}
\vspace{-0.5cm}
\caption{Comparison of the performance of CLIP, Full Fine-Tuning (FLYP), LoRA, and LoRA combined with our method across different ranks (4, 16, 32, 64, 128, and 256) with a lambda value of 1. The models are fine-tuned on ImageNet-1K and evaluated on zero-shot classification datasets. \textit{R} denotes the rank, and \textit{Blocks (\%)} indicates the percentage of activated blocks.}
\label{tab:lora-cls}
\end{table*}

\begin{table*}
\begin{center}
\begin{tabular*}{\linewidth}{@{\extracolsep{\fill}} lccccccccccccccccccccccc}
\toprule
\multirow{2}{*}{Model} & \multirow{2}{*}{R} & \multirow{2}{*}{Blocks (\%)} & \multirow{2}{*}{IN-1K} & \multicolumn{10}{c}{Zero-Shot Classification Datasets} & \multirow{2}{*}{Mean} \\
\cmidrule{5-14}
 & & & & C10 & C100 & DTD & MNI & SUN & AIR & FLW & FOD & Pets & Mean & \\
\toprule
CLIP & - & - &  68.3&90.8&68.3&44.7&59.4&65.3&24.3&71.0&88.5&89.1&66.8&67.6\\
\midrule
FLYP & - & - &  82.5&89.7&64.9&35.6&44.6&52.8&8.5&39.1&62.7&77.3&52.8&67.7\\
\midrule
\multirow{7}{*}{DoRA} 
&4&100&79.8&92.8&68.7&43.6&59.3&64.6&19.1&65.2&84.1&88.5&65.1&72.4\\
&8&100&80.1&92.7&68.3&43.8&59.4&64.4&18.2&64.4&83.6&88.3&64.8&72.5\\
&16&100&80.2&92.6&68.4&43.5&54.4&63.9&18.9&65.1&83.0&88.4&64.3&72.2\\
&32&100&80.5&92.4&68.0&43.4&56.2&63.6&18.6&65.3&82.5&88.5&64.3&72.4\\
&64&100&81.0&92.4&68.4&43.9&56.4&63.2&18.5&64.6&81.8&88.9&64.2&72.6\\
&128&100&81.4&92.1&68.7&42.5&52.5&62.1&17.7&63.3&80.1&89.0&63.1&72.3\\
&256&100&81.8&91.6&67.8&41.8&52.0&61.3&16.4&58.9&78.1&88.0&61.8&71.8\\
\midrule
\multirow{7}{*}{+Ours} 
&4&1.39&79.1&93.3&69.5&45.1&69.8&65.1&20.5&66.5&85.5&89.1&67.2&73.1\\
&8&2.08&79.5&92.6&69.3&47.1&60.2&64.0&19.7&66.3&84.1&88.8&65.8&72.6\\
&16&2.08&79.9&92.3&68.4&46.1&63.4&63.6&20.8&64.1&84.1&88.4&65.7&72.8\\
&32&2.78&80.3&92.1&68.5&46.1&62.9&63.3&19.6&64.3&82.6&88.1&65.3&72.8\\
&64&3.47&80.9&92.4&68.0&46.6&57.0&61.5&18.0&59.2&80.0&88.3&63.4&72.2\\
&128&3.47&81.5&92.0&67.1&45.3&52.2&60.1&15.8&58.6&77.6&87.9&61.8&71.6\\
&256&3.47&81.8&92.0&67.5&46.2&51.8&59.8&15.3&57.0&75.8&87.8&61.5&71.7\\
\bottomrule
\end{tabular*}
\end{center}
\vspace{-0.5cm}
\caption{Comparison of the performance of CLIP, Full Fine-Tuning (FLYP), DoRA, and DoRA combined with our method across different ranks (4, 16, 32, 64, 128, and 256) with a lambda value of 1. The models are fine-tuned on ImageNet-1K and evaluated on zero-shot classification datasets. \textit{R} denotes the rank, and \textit{Blocks (\%)} indicates the percentage of activated blocks.}
\label{tab:dora-cls}
\end{table*}

\begin{table*}
\begin{center}
\begin{tabular*}{\linewidth}{@{\extracolsep{\fill}} lcccccccccccccccccccccc}
\toprule
\multirow{2}{*}{Model} & \multirow{2}{*}{R} & \multirow{2}{*}{Blocks (\%)} & \multirow{2}{*}{IN-1K} & \multicolumn{4}{c}{MS-COCO} & \multicolumn{4}{c}{Flicker-30K} & \multirow{2}{*}{Mean} \\
\cmidrule{5-8} \cmidrule{9-12}
 & & & & IR@1 & IR@5& TR@1& TR@5&IR@1& IR@5& TR@1&TR@5 & \\
 \toprule
 CLIP &- &-& 68.3&51.7&76.7&32.7&57.7&81.2&96.2&63.7&86.3&68.3\\
\midrule

FLYP &- &-&82.5&29.5&52.1&19.0&39.5&59.5&84.7&43.7&70.5&66.2\\

\midrule
\multirow{7}{*}{LoRA} 
&4&100&80.0&47.1&71.2&29.2&53.6&76.4&93.7&59.8&84.5&72.2\\
&8&100&80.4&45.7&69.2&28.2&52.4&75.2&93.6&57.9&83.7&71.8\\
&16&100&80.8&43.8&68.3&27.4&51.6&74.8&92.9&57.1&82.7&71.6\\
&32&100&81.3&41.4&65.7&26.3&50.2&72.0&91.7&54.9&81.3&70.9\\
&64&100&81.7&37.6&62.4&23.8&46.6&69.4&89.7&50.9&78.0&69.5\\
&128&100&81.8&29.4&53.3&18.3&39.6&59.3&83.6&42.1&69.7&65.6\\
&256&100&81.6&16.1&33.5&10.7&26.8&35.0&62.4&26.2&50.7&57.1\\

\midrule
\multirow{7}{*}{+Ours} 
&4&2.08&79.2&50.2&74.1&32.7&57.8&80.0&80.0&64.9&86.6&72.5\\
&8&2.08&79.3&50.5&73.8&32.7&57.9&79.3&94.4&63.8&86.6&73.3\\
&16&2.78&79.7&48.7&72.2&31.8&56.5&77.8&94.2&62.6&85.6&73.0\\
&32&3.47&80.4&46.3&70.6&30.7&55.4&77.0&92.5&60.3&84.6&72.5\\
&64&4.86&81.0&43.3&67.7&28.7&53.2&74.6&91.9&57.7&83.3&71.8\\
&128&6.25&81.8&39.4&64.9&25.6&49.0&70.4&90.4&52.3&79.8&70.4\\
&256&5.56&82.3&34.7&58.9&22.2&44.8&64.7&87.1&47.3&74.5&68.3\\

\bottomrule
\end{tabular*}
\end{center}
\vspace{-0.5cm}
\caption{Comparison of the performance of CLIP, Full Fine-Tuning (FLYP), LoRA, and LoRA combined with our method across different ranks (4, 16, 32, 64, 128, and 256) with a lambda value of 1. The models are fine-tuned on ImageNet-1K and evaluated on zero-shot retrieval datasets. \textit{R} denotes the rank, and \textit{Blocks (\%)} indicates the percentage of activated blocks.}
\label{tab:lora-ret}
\end{table*}

\begin{table*}
\begin{center}
\begin{tabular*}{\linewidth}{@{\extracolsep{\fill}} lcccccccccccccccccccccc}
\toprule
\multirow{2}{*}{Model} & \multirow{2}{*}{R} & \multirow{2}{*}{Blocks (\%)} & \multirow{2}{*}{IN-1K} & \multicolumn{4}{c}{MS-COCO} & \multicolumn{4}{c}{Flicker-30K} & \multirow{2}{*}{Mean} \\
\cmidrule{5-8} \cmidrule{9-12}
 & & & & IR@1 & IR@5& TR@1& TR@5&IR@1& IR@5& TR@1&TR@5 & \\
  \toprule
 CLIP &- &-& 68.3&51.7&76.7&32.7&57.7&81.2&96.2&63.7&86.3&68.3\\
\midrule

FLYP &- &-&82.5&29.5&52.1&19.0&39.5&59.5&84.7&43.7&70.5&66.2\\
\midrule
\multirow{7}{*}{DoRA} 
&4&100&80.0&47.7&72.1&29.9&54.2&76.3&94.0&60.2&84.4&72.3\\
&8&100&80.4&46.7&71.6&29.1&53.5&75.8&93.5&59.5&83.9&72.2\\
&16&100&80.8&46.2&70.9&28.7&52.9&75.7&93.3&58.0&83.6&71.9\\
&32&100&81.3&45.4&69.7&28.1&52.3&75.4&92.8&57.5&83.2&71.8\\
&64&100&81.7&44.5&68.9&27.5&51.5&74.2&92.6&56.3&82.8&71.6\\
&128&100&81.8&42.5&67.0&26.3&50.3&74.3&92.0&54.6&81.9&71.3\\
&256&100&81.6&40.0&65.2&25.0&48.5&71.7&91.9&52.2&79.4&70.5\\
\midrule
\multirow{7}{*}{+Ours} 
&4&1.39&79.1&50.7&74.2&33.2&58.0&80.6&95.1&64.1&86.8&73.5\\
&8&2.08&79.5&49.4&72.8&32.3&56.9&79.3&94.7&62.9&85.9&73.1\\
&16&2.08&79.9&48.8&72.3&31.7&56.4&78.7&94.4&61.8&85.6&73.0\\
&32&2.78&80.3&46.0&70.6&30.7&55.4&76.6&92.2&59.8&84.8&72.4\\
&64&3.47&80.9&43.4&68.3&28.9&53.5&74.4&92.1&57.8&83.4&71.8\\
&128&3.47&81.5&41.2&66.0&27.3&51.7&72.5&91.0&55.6&81.8&71.2\\
&256&3.47&81.8&39.6&64.2&26.1&50.5&71.5&92.0&54.6&80.5&70.9\\
\bottomrule
\end{tabular*}
\end{center}
\vspace{-0.5cm}
\caption{Comparison of the performance of CLIP, Full Fine-Tuning (FLYP), DoRA, and DoRA combined with our method across different ranks (4, 16, 32, 64, 128, and 256) with a lambda value of 1. The models are fine-tuned on ImageNet-1K and evaluated on zero-shot retrieval datasets. \textit{R} denotes the rank, and \textit{Blocks (\%)} indicates the percentage of activated blocks.}
\label{tab:dora-ret}
\end{table*}

\end{document}